\def\year{2016}\relax
\documentclass[letterpaper]{article}
\usepackage{aaai16}
\usepackage{times}
\usepackage{helvet}
\usepackage{courier}
\usepackage{graphicx}
\usepackage{bm}
\usepackage{multirow}
\usepackage{multicol}
\usepackage{amsmath}
\usepackage{epstopdf}
\usepackage{amsfonts,amssymb}
\usepackage{eqlist}
\usepackage{hhline}
\begin{document}
\title{Implicit Discourse Relation Classification via Multi-Task Neural Networks}

\author{Yang Liu$^1$, Sujian Li$^{1,2}$, Xiaodong Zhang$^{1}$ \and Zhifang Sui$^{1,2}$\\
$^1$ Key Laboratory of Computational Linguistics, Peking University, MOE, China\\
$^2$ Collaborative Innovation Center for Language Ability, Xuzhou, Jiangsu, China\\
{\tt \{cs-ly, lisujian, zxdcs, szf\}@pku.edu.cn }}
\maketitle
\begin{abstract}
\begin{quote}
Without discourse connectives, classifying implicit discourse relations is a challenging task and a bottleneck for building a practical discourse parser.
Previous research usually makes use of one kind of discourse framework such as PDTB or RST to improve the classification performance on discourse relations.
Actually, under different discourse annotation frameworks, there exist multiple corpora which have internal connections.
To exploit the combination of different discourse corpora, we design related discourse classification tasks specific to a corpus,
and propose a  novel Convolutional Neural Network embedded multi-task learning system to synthesize these tasks by learning both unique and  shared representations for each task.
The experimental results on the PDTB implicit discourse relation classification task demonstrate that our model achieves significant gains over  baseline systems.\end{quote}
\end{abstract}

\section{Introduction}
Discourse relations (e.g., contrast and causality) support a set of sentences to form a coherent text.
Automatically identifying these discourse relations can help many downstream NLP tasks such as question answering and automatic summarization.

Under certain circumstances, these relations are in the form of explicit markers like ``but'' or ``because'', which is relatively easy to identify.
Prior work~\cite{pitler2009automatic} shows that where explicit markers exist, relation types  can be disambiguated with $F_1$ scores higher than 90\%.
However, without an explicit marker to rely on, classifying the  implicit discourse relations  is much more difficult.
The fact that these implicit  relations  outnumber the explicit ones in naturally occurring text makes the classification of their  types   a key challenge in  discourse analysis.

The major line of research work approaches the implicit relation classification problem by extracting informed features from the corpus and designing  machine learning algorithms~\cite{pitler2009automatic,lin2009recognizing,louis2010using}. An obvious challenge for classifying discourse relations is which features are appropriate for representing the sentence pairs. Intuitively, the word pairs occurring in the sentence pairs are useful, since they to some extent can represent semantic relationships between two sentences (for example, word pairs  appearing around the contrast relation often tend to be antonyms).
In earlier studies,  researchers find that word pairs  do help classifying the discourse relations. However, it is strange that most of these useful word pairs are composed of  stopwords.  \citeauthor{rutherford-xue:2014:EACL} (2014) point out that this counter-intuitive phenomenon is caused by the sparsity nature of these word pairs.
They employ Brown clusters as an alternative abstract word representation, and as a result, they get more  intuitive cluster pairs  and achieve a better performance.

Another problem in discourse parsing is the coexistence of different discourse annotation frameworks, under which different kinds of corpora and tasks are created. The well-known discourse corpora include the Penn Discourse TreeBank (PDTB)~\cite{prasad2007penn} and the Rhetorical Structure Theory - Discourse Treebank (RST-DT)~\cite{mann1988rhetorical}. Due to the annotation complexity, the size of each corpus is not large enough. Further, these corpora under different annotation frameworks are usually used separately in discourse relation classification,  which is also the main reason of sparsity in discourse relation classification.
However, these different annotation frameworks have strong internal connections.
For example, both the \textit{Elaboration} and \textit{Joint} relations in RST-DT  have a similar sense as the \textit{Expansion} relation in PDTB.
Based on this, we consider to design multiple discourse analysis tasks according to these frameworks and synthesize these tasks with the goal of classifying the implicit discourse relations, finding more precise representations for the sentence pairs.
The most inspired work to us is done by \cite{lan2013leveraging} where they regard implicit and explicit relation classification in PDTB framework as two tasks and design a multi-task learning method to obtain higher  performance.

In this paper, we propose a more general multi-task learning system for implicit discourse relation classification by synthesizing the discourse analysis tasks within different corpora.
To represent the sentence pairs, we construct the convolutional neural networks (CNNs) to derive their  vector representations in a low dimensional latent space, replacing the sparse lexical features.
To combine different  discourse analysis tasks, we further embed the CNNs into a multi-task neural network and learn both the unique and shared representations for the sentence pairs in different tasks, which can reflect the differences and connections among these tasks.
With our multi-task neural network, multiple discourse tasks are trained simultaneously and optimize each other through their connections.

\begin{table*}
  \small
  \center
\begin{tabular}{|l|l|p{5.8cm}|p{5.8cm}|}
\hline
Data Source&Discourse Relation&Argument 1&Argument 2\\
\hline
RST-DT&Elaboration&it added 850 million Canadian dollars&reserves now amount to 61\% of its total less-developed-country exposure\\
\hline
PDTB&Expansion(implicit)&Income from continuing operations was up 26\%&Revenue rose 24\% to \$6.5 billion from \$5.23 billion\\
\hline
PDTB&Expansion(explicit)&as in summarily sacking exchange controls&and \textit{in particular} slashing the top rate of income taxation to 40\%\\
\hline
NYT Corpus&particularly&Native plants seem to have a built-in tolerance to climate extremes and have thus far done well&\textit{particularly}  fine show-offs have been the butterfly weeds, boneset and baptisia\\
\hline
\end{tabular}
\caption{Discourse Relation Examples in Different Corpora. }
\label{Tab:rel}
\end{table*}

\section{Prerequisite }
As stated above, to improve the implicit discourse relation classification, we make full use of the combination of different discourse corpora.
In our work, we choose three kinds of discourse corpora: PDTB, RST-DT and the natural text with discourse connective words.
In this section, we briefly introduce the corpora.

\subsubsection{PDTB}
The Penn Discourse Treebank (PDTB)~\cite{prasad2007penn}, which is known as the largest discourse corpus, is composed of 2159 Wall Street Journal articles. PDTB adopts the predicate-argument structure, where the predicate is the discourse connective (e.g. while) and the arguments  are two text spans around the connective.
In PDTB, a relation is explicit if there is an explicit discourse connective presented in the text; otherwise, it is implicit.
All  PDTB relations are hierarchically organized  into 4 top-level classes: \textit{Expansion}, \textit{Comparison}, \textit{Contingency}, and \textit{Temporal} and can be further divided into 16 types and 23 subtypes. In our work, we mainly experiment on the 4 top-level classes as in previous work~\cite{lin2009recognizing}.

\subsubsection{RST-DT}
RST-DT is based on the Rhetorical Structure Theory (RST) proposed by~\cite{mann1988rhetorical} and is composed of 385 articles. In this corpus, a  text is represented as a discourse tree whose leaves are non-overlapping text spans called elementary discourse units (EDUs).
Since we mainly focus on discourse relation classification, we make use of the discourse dependency structure~\cite{li2014text} converted from the tree structures and extracted the EDU pairs with labeled rhetorical relations between them.
In RST-DT, all relations are classified into 18 classes. We choose the highest frequent 12 classes and get 19,681 relations.

\subsubsection{Raw Text with Connective Words}
There exists a large amount of raw text with connective words in it.
As we know, these connective words serve as a natural means to connect text spans.
Thus, the raw text with connective words is somewhat similar to the explicit discourse relations in PDTB without expert judgment and can  also be used as a special discourse corpus.
In our work, we adopt the New York Times (NYT) Corpus~\cite{sandhaus2008new} with over 1.8 million news articles. We extract the sentence pairs around the 35 commonly-used connective words, and generate a new discourse corpus with  40,000 relations after removing the connective words.
This corpus is not verified by human and contains some noise, since not all the connective words reflect  discourse relations or some connective words may have different meanings in different contexts.
However,  it can still help training a better model with a certain scale of instances.

\section{Multi-Task Neural Network for Discourse  Parsing}

\subsection{Motivation and Overview}
Different discourse corpora are closely related, though under different annotation theories.
In Table \ref{Tab:rel}, we list some instances which have similar discourse relations in nature but are annotated differently in different corpora.
The second row belongs to the \textit{Elaboration} relation in RST-DT.
The third and fourth row are both \textit{Expansion} relations in PDTB: one is implicit and the other is explicit with the connective ``\textit{in particular}''.
The fifth row is from the NYT corpus and directly uses the word ``\textit{particularly}'' to denote the discourse relation between two sentences.
From these instances, we can see they all reflect the similar discourse relation  that the second argument gives more details of the first argument.
It is intuitive that the  classification performance on these instances can be boosted from each other if we  appropriately synthesize them.
With this idea, we propose to adopt the multi-task learning method and design a specific discourse analysis task for each corpus.

According to the principle of multi-task learning, the more related the tasks are, the more powerful the multi-task learning method will be. Based on this, we design four discourse relation classification tasks.
\begin{description}
\item[Task 1] Implicit PDTB Discourse Relation Classification
\item[Task 2] Explicit PDTB Discourse Relation Classification
\item[Task 3] RST-DT Discourse Relation Classification
\item[Task 4] Connective Word Classification
\end{description}

The first two tasks  both classify the  relation between two arguments in  the PDTB framework.
The third task is to predict the relations between two EDUs using our processed RST-DT corpus.
The last one is designed to predict a correct connective word to a sentence pair by using the NYT  corpus.
We define the task of classifying implicit PDTB relations as our main task, and the other tasks as the auxiliary tasks.
This means we will focus on learning from other tasks to improve the performance of our main task in our system.
It is noted that we call the two text spans in all the tasks as arguments for convenience.

Next, we will introduce how we  tackle these tasks.
In our work, we propose to use the convolutional neural networks (CNNs) for representing the argument pairs.
Then, we embed the CNNs into a multi-task neural network (MTNN), which can learn the shared and unique properties of all the tasks.

\subsection{CNNs: Modeling Argument Pairs}

\begin{figure}[!htbp]
\centering
\includegraphics[width=3.2in]{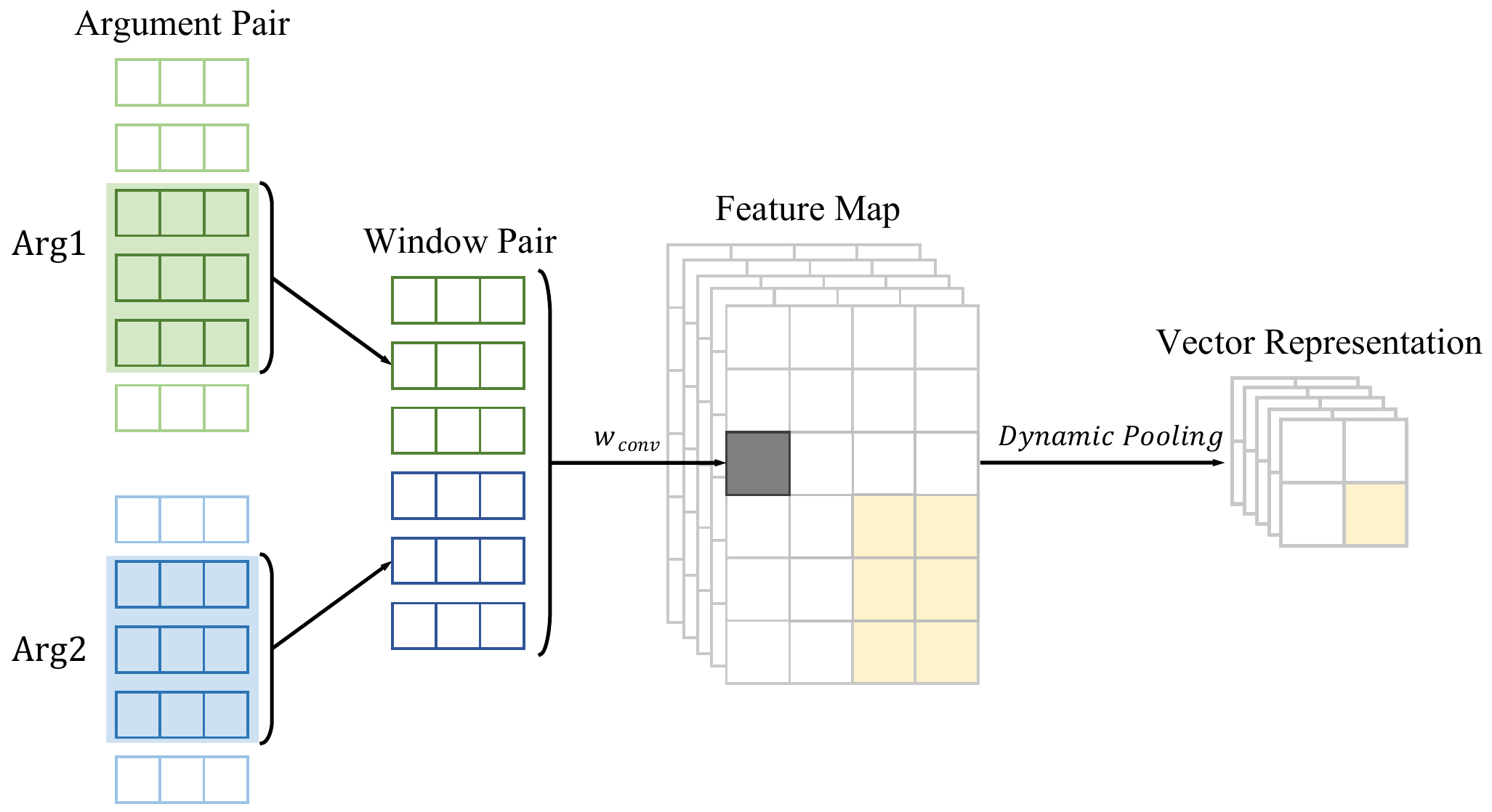}
\caption{Neural Networks For Modeling the Argument Pair. }
\label{fig:conv}
\end{figure}

Figure \ref{fig:conv}  illustrates our proposed method of modeling the argument pairs.
We associate each word $w$  with a vector representation $\bm{x}_w \in \mathbb{R}^{D_e}$, which is usually pre-trained with large unlabeled corpora.
We view an argument as a sequence of these word vectors, and let $\bm{x}_i^1$ ($\bm{x}_i^2$) be the  vector of the $i$-th word in  argument $Arg1$ ($Arg2$).
Then, the argument pair can be represented as,
\begin{gather}
Arg1: [\bm{x}^1_1, \bm{x}^1_2, \cdots, \bm{x}^1_{m_1}]\\
Arg2: [\bm{x}^2_1, \bm{x}^2_2, \cdots, \bm{x}^2_{m_2}]
\end{gather}
where $Arg1$ has $m_1$ words and $Arg2$ has $m_2$ words.

Generally, let $\bm{x}_{i:i+j}$ relate to the concatenation of word vectors $\bm{x}_i, \bm{x}_{i+1}, \cdots, \bm{x}_{i+j}$. A convolution operation involves a filter $\bm{w}$, which is applied to a window of $h$ words to produce a new feature.
For our specific task of capturing the relation between two arguments, each time we take $h$ words from each arguments, concatenate their vectors, and apply the convolution operation on this window pair. For example, a feature $c_{ij}$ is generated from a window pair composed by words $\bm{x}^1_{i:i+h-1}$ from $Arg1$ and words $\bm{x}^2_{j:j+h-1}$  from $Arg2$,
\begin{gather}
c_{ij} = f(\bm{w}\cdot [\bm{x}^1_{i:i+h-1}, \bm{x}^2_{j:j+h-1}] + b)
\end{gather}
where $b$ is a bias term and $f$ is a non-linear
function, for which in this paper we use $tanh$.

The filter is applied to each possible window pair of the two arguments to produce a  feature map $\bm{c}$, which is a two-dimensional matrix. Since the arguments may have different lengths, we use an operation called ``dynamic pooling'' to capture the most salient features in $\bm{c}$, generating a fixed-size  matrix $\bm{p} \in \mathbb{R}^{n_p\times n_p}$.
In order to do this, matrix $\bm{c}$ will be divided
into $n_p$ roughly equal parts. Every maximal value in the
rectangular window is selected to form a $n_p\times n_p$ grid.
During this process, the matrix $\bm{p}$ will lose some information compared to the original matrix  $\bm{c}$.
However,  this approach can capture $\bm{c}$'s global structure. For example, the upper left part of $\bm{p}$ will be constituted by word pair features reflecting the relationship between the beginnings of two arguments.
This property is useful to discourse parsing, because some prior research~\cite{lin2009recognizing} has pointed out that word position in one argument is  important for identifying the discourse relation.

With multiple filters like this,  the argument pairs can be modeled as a three-dimensional tensor. We flatten it to a vector $\bm{p} \in \mathbb{R}^{n_p\times n_p\times n_f}$ and use it as the representation of the argument pair, where $n_f$ is the number of filters.

\subsection{Multi-Task Neural Networks: Classifying Discourse Relations}
\begin{figure}[!htbp]
	\centering
	\includegraphics[height=2in]{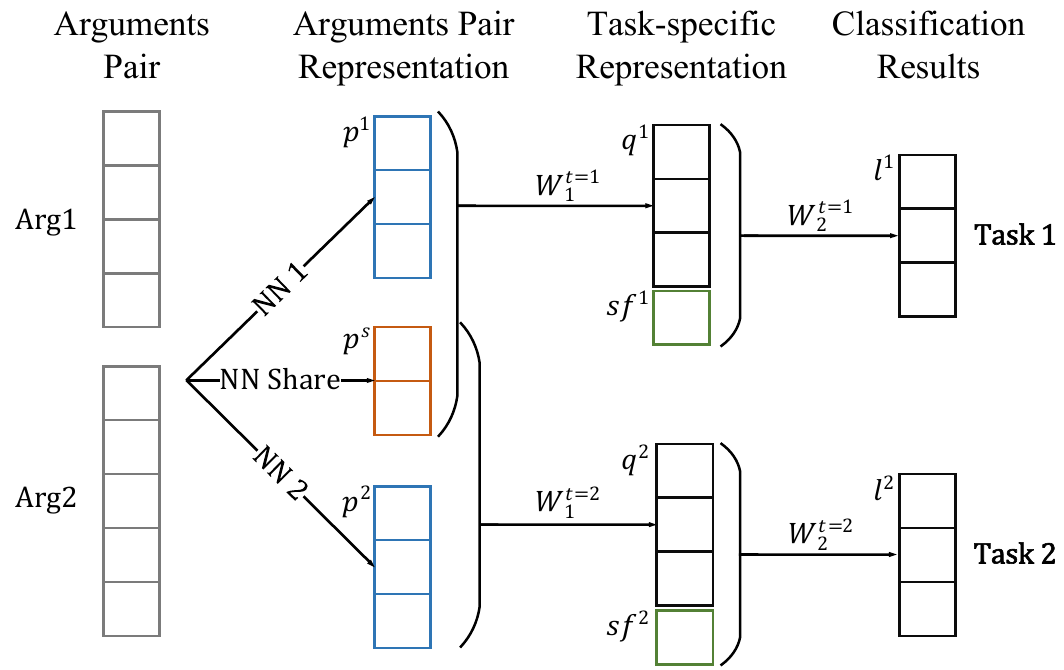}
	\caption{ Architecture of Multi-task Neural Networks for Discourse Relation Classification.}
	\label{fig:arc}
\end{figure}

Multi-task learning (MTL) is a kind of machine learning approach, which trains both the main task and auxiliary tasks simultaneously with a shared representation learning the commonality among the tasks.
In our work, we embed the convolutional neural networks into a multi-task learning system to synthesize the four tasks mentioned above.
We map the argument pairs of different tasks into  low-dimensional vector representations with the proposed CNN.
To guarantee  the principle that these tasks can optimize each other without bringing much noise, each task owns a unique representation of the argument pairs, meanwhile, there is  a special shared representation connecting all the tasks.
The architecture of our multi-task learning system is shown in Figure~\ref{fig:arc}.
For clarity, the diagram  depicts  only   two tasks.  It should be aware that the number of tasks  is not limited to two.

For task $t$, the argument pair $s = (Arg1, Arg2)$ is mapped into a unique vector $\bm{p}^t$ and a shared vector $\bm{p}^{s}$, where $NN$ denotes the convolutional neural networks for modeling the argument pair,
\begin{gather}
\bm{p}^t = NN_t(Arg1, Arg2)\\
\bm{p}^{s} = NN(Arg1, Arg2)\
\end{gather}

These two vectors are then concatenated and mapped into a task-specific representation $\bm{q}^t$ by a nonlinear transformation,
\begin{gather}
\bm{q}^t = f(\bm{w}^t_1[\bm{p}^t, \bm{p}^{s}]+ \bm{b}^t_1)
\end{gather}
where $\bm{w}^t$ is the transformation matrix and $\bm{b}^t_1$ is the bias term.

After acquiring  $\bm{q}^t$, we use several additional surface-level features, which have been proven useful in a bunch of existing work~\cite{lin2009recognizing,rutherford-xue:2014:EACL}. We notate the feature vector for task $t$ as $\bm{sf}^t$. Then, we concatenate $\bm{sf}^t$ with $\bm{q}^t$ and name it $\bm{r}^t$.
Since all  the  tasks are related with classification, we set the dimension of the output vector for task $t$ as the predefined class number $n_t$.
Next, we take $\bm{r}^t$ as input and generate the output vector $\bm{l}^t$ through a $softmax$ operation with the weight matrix $\bm{w}_t^2$ and the bias $\bm{b}^t_2$,
\begin{gather}
\bm{r}^t = [\bm{q}^t,\bm{sf}^t]\\
\bm{l}^t = softmax( \bm{w}^t_2\bm{r}^t + \bm{b}^t_2 )
\end{gather}
where the $i$-th dimension of  $\bm{l}^t$  can be interpreted as the conditional probability that an instance belongs to class $i$ in task $t$.

This network architecture has various good properties.
The shared representation makes sure these tasks can effectively learn from each other.
Meanwhile, multiple CNNs for modeling the argument pairs give us the flexibility to assign different hyper-parameters to each task. For example, PDTB is built on sentences while RST-DT is on elementary discourse units, which are usually shorter than sentences.
Under the proposed framework, we can assign a larger window size to PDTB related tasks and a smaller window size to the RST-DT related task, for better capturing their discourse relations.

\subsubsection{Additional Features}
When classifying the  discourse relations, we  consider several surface-level features, which are supplemental to the automatically generated representations.
We use different features for each task, considering their specific properties. These features include:
\begin{itemize}
	\item The first and last words of the arguments (For task 1)
	\item Production rules extracted from the constituent parse trees of the arguments (For task 1,2,4)
	\item Whether two EDUs are in the same sentence (For task 3)
\end{itemize}

\subsection{Model Training}

We define the ground-truth label vector $\bm{g}^t$ for each
instance in task $t$ as a binary vector. If the instance belongs to  class $i$, only the $i$-th dimension $\bm{g}^t[i]$ is 1 and the other dimensions are set to 0.
In our MTNN model, all the tasks are classification problems and we adopt cross entropy loss as the optimization function.
Given  the neural network parameters  $\Theta$ and the word embeddings $\Theta_e$, the objective function for instance $s$ can be written as,
\begin{equation}
J(\Theta, \Theta_e) = -\sum_i^{n_t}\bm{g}^t[i]\bm{l}^t[i]
\end{equation}

We use mini-batch stochastic gradient descent (SGD) to train the parameters $\Theta$ and $\Theta_e$.
Referring to the training procedure in~\cite{liu2015representation}, we select one task in each epoch and update the model according to its task-specific objective.

To avoid over-fitting, we use different learning rates to train the neural network parameters and the word embeddings , which are denoted as $\lambda$ and $\lambda_e$.
To make the most of all the tasks, we expect them to reach their best performance at roughly the same time. In order to achieve this, we assign different regulative ratio $\mu$ and $\mu_e$ to different tasks, adjusting their learning rates $\lambda$ and $\lambda_e$. That is, for task $t$, the update rules for  $\Theta$ and $\Theta_e$ are,
\begin{gather}
\Theta \leftarrow \Theta + \mu^t\lambda\frac{\partial J(\Theta)}{\partial\Theta}\\
\Theta_e \leftarrow \Theta_e + \mu_e^t\lambda_e\frac{\partial J(\Theta_e)}{\partial\Theta_e}
\end{gather}
It is worth noting that, to avoid bringing noise to the main task, we let  $\mu_e$ of the auxiliary tasks  to be very small, preventing them from changing the word embeddings too much.

\section{Experiments}
\subsection{Datasets}
As introduced above,  we use three corpora:  PDTB, RST-DT, and the NYT corpus, in our experiments to train our multi-task neural network.

\begin{table}[!htbp]
\center
  \begin{tabular}{|l|c|c|c|}
  \hline
Relation&Train&Dev&Test\\
\hline
Comparison&1855&189&145\\
Contingency&3235&281&273\\
Expansion&6673&638&538\\
Temporal&582&48&55\\
\hline
\textbf{Total}&12345&1156&1011\\
  \hline
\end{tabular}
\caption{Distribution of Implicit Discourse Relations in  PDTB.}
  \label{tab:imp}
\end{table}

Since our main goal is to conduct implicit discourse relation classification (the main task), Table~\ref{tab:imp} summarizes the statistics of the four top-level implicit discourse relations in  PDTB.
We follow the setup of previous studies~\cite{pitler2009automatic}, splitting the dataset into a a training set, development set, and test set.  Sections 2-20 are used to train classifiers, Sections 0-1  to develop feature sets and tune models, and Section 21-22 to  test the systems.

\begin{table}[!htbp]
\center
  \begin{tabular}{|l|c|l|c|}
\hhline{|--|--|}
  Relation&Freq.&Relation&Freq.\\
\hhline{|--|--|}
Comparison&5397&Temporal&2925\\
Contingency&3104&Expansion&6043\\
\hhline{|--|--|}
\end{tabular}
\caption{Distribution of Explicit Discourse Relations in  PDTB.}
  \label{tab:pdtb-exp}
\end{table}

For Task 2,  all 17,469 explicit relations in sections 0-24 in PDTB are used. Table~\ref{tab:pdtb-exp} shows the distribution of these explicit relations on four  classes.
For Task 3, we convert RST-DT trees to discourse dependency trees according to~\cite{li2014text} and  get  direct relations between EDUs, which is more suitable for the classification task. We choose the 12 most frequent coarse-grained relations shown in Table~\ref{tab:rst}, generating a corpus with 19,681 instances.

\begin{table}[!htbp]
\center
  \begin{tabular}{|l|c|l|c|}
\hhline{|--|--|}
  Relation&Freq.&Relation&Freq.\\
\hhline{|--|--|}
Elaboration&7675&Background&897\\
Attribution&2984&Cause&867\\
Joint&1923&Evaluation&582\\
Same-unit&1357&Enablement&546\\
Contrast&1090&Temporal&499\\
Explanation&959&Comparison&302\\
\hhline{|--|--|}
\end{tabular}
\caption{ Distribution of 12 Relations Used in RST-DT.}
  \label{tab:rst}
\end{table}
For Task 4, we use the Standford parser \cite{klein2003accurate} to segment sentences.
We select the 35 most frequent connectives in PDTB, and extract instances containing these connectives from the NYT corpus based on the same patterns as in~\cite{rutherford-xue:2015:NAACL-HLT}. We then manually compile a set of rules to remove some noisy instances, such as those with too short arguments.
Finally, we obtain a corpus with 40,000 instances by random sampling. Due to space limitation, we only list the 10 most frequent connective words in our corpus in Table~\ref{tab:nyt}.
\begin{table}[!htbp]
\center
  \begin{tabular}{|l|c|l|c|}
\hhline{|--|--|}
  Relation&Pct.&Relation&Pct.\\
\hhline{|--|--|}
Because&22.52\%&For example&5.92\%\\
If&8.65\%&As a result&4.30\%\\
Until&9.45\%&So&3.26\%\\
In fact&9.25\%&Unless&2.69\%\\
Indeed&8.02\%&In the end&2.59\%\\
\hhline{|--|--|}
\end{tabular}
\caption{ Percentage of 10 Frequent Connective Words Used in NYT Corpus.}
  \label{tab:nyt}
\end{table}

\subsection{Model Configuration}
We use word embeddings provided by GloVe~\cite{pennington2014glove}, and the dimension of the embeddings $D_e$ is 50.
We first train these four tasks separately to roughly set their hyper-parameters. Then, we more carefully tune the multi-task learning system based on the performance of our main task on the development set.
The learning rates are set as $\lambda = 0.004, \lambda_e = 0.001$.

Each task has a set of hyper-parameters, including  the window size of CNN $h$, the pooling size $n_p$, the number of filters $n_f$, dimension of the task-specific representation $n_r$, and the regulative ratios  $\mu$ and $ \mu_e$. All the tasks share a window size, a pooling size and a number of filters for learning the shared representation, which are denoted as $h^s, n^s_p, n^s_f$. The detailed settings are shown in Table~\ref{tab:hyper}.
\begin{table}[!htbp]
	\small
	\center
	\begin{tabular}{|c|c|c|c|c|c|c|c|c|c|}
		\hline
		{Task} &$h$&$n_p$&$n_f$&$n_r$&$\mu$&$\mu_e$&$h^s$&$n^s_p$&$n^s_f$\\
		\hline
		1&5&10&80&20&1.0&1.0&\multirow{4}{*}{6}&\multirow{4}{*}{10}&\multirow{4}{*}{40}\\
		2&5&10&80&20&1.5&0.15&&&\\
		3&4&8&100&30&2.0&0.2&&&\\
		4&4&10&100&40&2.0&0.2&&&\\
		\hline
	\end{tabular}
	\caption{Hyper-parameters for the MTL system.}\label{tab:hyper}
\end{table}

\subsection{Evaluation and Analysis}
We mainly evaluate the performance of the implicit PDTB relation classification, which can be seen as a 4-way classification task.
For each relation class, we adopt the commonly used metrics,  Precision, Recall and $F_1$, for performance evaluation.
To evaluate the whole system, we use the metrics of Accuracy and macro-averaged $F1$.

\subsubsection{Analysis of Our Model}
First of all, we evaluate the combination of different tasks.
Table \ref{tab:results} shows the detailed results.
For each relation, we first conduct the main task (denoted as 1) through implementing a CNN model and show the results in the first row.
Then we combine the main task with one of the other three auxiliary tasks (i.e., 1+2, 1+3, 1+4) and  their results in the next three rows.
The final row gives the performance using all the four tasks (namely, ALL). In general, we can see that when synthesizing all the tasks, our MTL system can achieve the best performance.

\begin{table}[!htbp]
	\center
	\small
	\begin{tabular}{|l|l|>{\centering\arraybackslash}p{3.7em}|>{\centering\arraybackslash}p{3.7em}|>{\centering\arraybackslash}p{3.7em}|}
		\hline
		{Relation} &
		{Tasks} &
		Precision &
		Recall&$F_1$
		\\
		\hline
		\multirow{3}{*}{Expansion}
&1&59.47&74.72&66.23\\
&1+2&60.64&71.00&65.41\\
&1+3&60.35&71.56&65.48\\
&1+4&60.00&\textbf{77.51}&67.64\\
&ALL&\textbf{64.13}&76.77&\textbf{69.88}\\

		\hline
		\multirow{3}{*}{Comparison}
&1&34.65&30.35&32.35\\
&1+2&30.00&22.76&25.88\\
&1+3&\textbf{40.37}&30.34&\textbf{34.65}\\
&1+4&35.94&15.86&22.01\\
&ALL&30.63&\textbf{33.79}&32.13\\
		\hline
		\multirow{3}{*}{Temporal}
&1&35.29&10.91&16.67\\
&1+2&36.36&21.82&27.27\\
&1+3&37.50&16.36&22.79\\
&1+4&\textbf{60.00}&10.91&18.46\\
&ALL&42.42&\textbf{25.45}&\textbf{31.82}\\
		\hline
		\multirow{3}{*}{Contingency}
&1&42.93&30.04&35.35\\
&1+2&40.34&35.17&37.57\\
&1+3&42.50&37.36&39.77\\
&1+4&47.11&\textbf{41.76}&44.27\\
&ALL&\textbf{59.20}&37.73&\textbf{46.09}\\
		\hline
	\end{tabular}
	\caption{Results on  4-way Classification of Implicit Relations in PDTB. }\label{tab:results}
\end{table}

More specifically, we find these  tasks have different influence on different discourse relations.
Task 2, the classification of explicit PDTB relations, has slight or even negative impact on the relations  except the \textit{Temporal} relation. This result is consistent with the conclusion reported in~\cite{sporleder2008using}, that there exists difference between explicit and implicit discourse   relations and more corpus of explicit relations does not definitely boost the performance of implicit ones.
Task 4, the classification of connective words, besides having the similar effects, is observed to be greatly helpful for identifying the \textit{Contingency} relation.
This may be because the \textit{Contingency} covers a wide range of subtypes and the fine-grained connective words in NYT corpus can give some hints of identifying this relation.
On the contrary, when training with the task of classifying RST-DT relations (Task 3), the result gets better on \textit{Comparison}, however, the improvement on other relations is less obvious than when using the other two tasks.
One possible reason for this is the definitions of \textit{Contrast} and \textit{Comparison} in RST-DT are similar to \textit{Comparison} in PDTB, so these two tasks can more easily learn from each other on these classes.
Importantly, when synthesizing all the tasks in our model, except the result on \textit{Comparison} relation experiences a slight deterioration, the classification performance generally  gets better.
\subsubsection{Comparison with Other Systems}
\begin{table}[!htbp]
	\center
	\begin{tabular}{|c|c|c|}
\hline
    System&Accuracy&$F_1$
    \\
    \hline
    \cite{rutherford-xue:2015:NAACL-HLT}&57.10&40.50\\
    Proposed STL&52.82&37.65\\
    Proposed MTL&\textbf{57.27}&\textbf{44.98}\\
    \hline
    \end{tabular}
    \caption{ General Performances of Different Approaches on  4-way Classification Task. }\label{tab:results2}
\end{table}

We  compare the general performance of our model with a state-of-the-art system in terms of accuracy and macro-average $F_1$ in Table \ref{tab:results2}.
\citeauthor{rutherford-xue:2015:NAACL-HLT} (2015) elaborately select a  combination of various lexical features, production rules, and Brown cluster pairs, feeding them into a maximum entropy classifier.
They also propose to gather weakly labeled data based on the discourse connectives for the classifier and achieve state-of-the-art results on 4-way classification task. We can see our proposed MTL system achieves higher performance on both accuracy and macro-averaged $F_1$.
We also compare the general performance between our MTL system and the Single-task Learning (STL) system which is only trained on Task 1.
 The result shows MTL raises the Accuracy from 52.82 to 57.27 and the  $F1$  from 37.65 to 44.98. Both improvements are significant under one-tailed t-test ($p<0.05$).

\begin{table}[!htbp]
\small
	\center
	\begin{tabular}{|c|c|c|c|c|}
\hline
	System&Comp.&Cont.&Expa.&Temp.\\\hline
	\cite{39260331}&31.79&47.16&-&20.30\\
	 \cite{park2012improving}&31.32&49.82&-&26.57\\
     \cite{TACL536}&35.93&52.78&-&27.63
    \\
    (R\&X 2015)&\textbf{41.00}&53.80&69.40&33.30\\
    Proposed STL&37.10&51.73&67.53&29.38\\
    Proposed MTL&37.91&\textbf{55.88}&\textbf{69.97}&\textbf{37.17}\\
    \hline
    \end{tabular}
    \caption{  General Performances of Different Approaches on Binary Classification Task.}\label{tab:results3}
\end{table}

For a more direct comparison with previous results,
we also conduct experiments based on the setting
that the task as four binary one vs. other classifiers. The
results are presented in Table~\ref{tab:results3}.
Three additional systems are used as baselines.
 \citeauthor{park2012improving} (2012) design a traditional feature-based method and promote the performance through optimizing the feature set.
\citeauthor{TACL536} \shortcite{TACL536} used two recursive neural networks on the syntactic parse tree to induce the representation of the arguments  and  the entity spans.
\citeauthor{39260331} \shortcite{39260331} first predict connective words on a unlabeled corpus, and then use these these predicted connectives as features to recognize the discourse relations.

The results show that the multi-task learning system is especially helpful for classifying the \textit{Contingency} and \textit{Temporary} relation. It increases the performance on \textit{Temporary} relation from 33.30 to 37.17, achieving a substantial improvement. This is probably because this relation suffers from the lack of training data in STL, and the use of MTL can learn  better representations for argument pairs, with the help of  auxiliary tasks.
The \textit{Comparison} relation benefits the least from MTL. Previous work of~\cite{rutherford-xue:2014:EACL} suggests this relation relies on the syntactic information of two arguments. Such features are captured in the upper layer of our model, which can not be optimized by multiple tasks.
Generally, our system achieves the state-of-the-art performance on three discourse relations (\textit{Expansion}, \textit{Contingency} and \textit{Temporary}).

\section{Related Work}
The supervised method often approaches discourse analysis as a classification problem of pairs of sentences/arguments. The first work to tackle this task on PDTB were~\cite{pitler2009automatic}. They selected several surface features to train four binary classifiers, each for one of the top-level PDTB relation classes.
Although other features proved to be useful, word pairs were  the major contributor to most of these classifiers.
Interestingly, they found that training these features on PDTB was more useful than training them on an external corpus.
Extending from this work, \citeauthor{lin2009recognizing} (2009) further identified four different feature types representing the context, the constituent parse trees, the dependency parse trees and the raw text respectively.
In addition, \citeauthor{park2012improving} (2012) promoted the performance through optimizing the feature set.
Recently, \citeauthor{mckeown2013aggregated} (2013) tried to tackle the feature sparsity problem by aggregating features.
\citeauthor{rutherford-xue:2014:EACL} (2014) used brown cluster to replace the word pair features, achieving the state-of-the-art classification performance.
\citeauthor{TACL536} \shortcite{TACL536} used two recursive neural networks  to represent the  arguments  and  the entity spans and use the combination of the representations to predict the discourse relation.

There also exist some semi-supervised approaches which exploit both labeled and unlabeled data for discourse relation classification.
\citeauthor{39286423} (2010) proposed a semi-supervised method to exploit the co-occurrence of features in unlabeled data. They found this method was  especially effective for improving  accuracy for infrequent relation types.
\citeauthor{39260331} \shortcite{39260331} presented a method to predict the missing connective based on a language model trained on an unannotated corpus. The predicted connective was then used as a feature to classify the implicit relation.
An interesting work is done by~\cite{lan2013leveraging}, where they designed a multi-task learning method to improve the classification performance by leveraging both implicit and explicit discourse data.

In recent years, neural network-based methods have gained prominence in the field of natural language processing~\cite{kim:2014:EMNLP2014,cao2015ranking}.
Some multi-task neural networks are proposed. For example, \citeauthor{collobert2011natural} (2011) designed a single sequence labeler for multiple tasks, such as Part-of-Speech tagging, chunking, and named entity recognition. Very recently, \citeauthor{liu2015representation} \shortcite{liu2015representation}  proposed a  representation learning algorithm based on multi-task objectives, successfully combining the tasks of query classification and web search.

\section{Conclusion}
Previous studies on implicit discourse relation classification always face two problems:  sparsity and argument representation.
To solve these two problems, we propose to use different kinds of corpus and design a multi-task neural network (MTNN) to synthesize different corpus-specific discourse classification tasks.
In our MTNN model, the convolutional neural networks with dynamic pooling are developed to model the argument pairs.
Then, different discourse classification tasks can derive their unique and shared representations for the argument pairs, through which they can optimize each other without bringing useless noise.
Experiment results demonstrate that our system achieves state-of-the-art performance.
In our future work, we will design a MTL system based on the syntactic tree, enabling each task to share the structural information.

\section{ Acknowledgments}
We thank all the anonymous reviewers for their insightful comments on this paper.
This work was partially supported by National Key Basic Research Program of China (2014CB340504),
National Natural Science Foundation of China (61273278 and 61572049).
The correspondence author of this paper is Sujian Li.

\bibliographystyle{aaai}
\bibliography{aaai}
\end{document}